%% file: icme2023template.tex
\documentclass{article}
\usepackage{spconf,amsmath,epsfig}
\usepackage{booktabs} 
\usepackage{color,xcolor}

\let\OLDthebibliography\thebibliography
\renewcommand\thebibliography[1]{
  \OLDthebibliography{#1}
  \setlength{\parskip}{0pt}
  \setlength{\itemsep}{0pt plus 0.3ex}
}

\newcommand{\modelname}{COM3D}

\begin{document}\sloppy

% Example definitions.
% --------------------
\def\x{{\mathbf x}}
\def\L{{\cal L}}

% \setlength{\abovecaptionskip}{0pt}

% Title.
% ------
\title{COM3D: Leveraging Cross-View Correspondence \\ and Cross-Modal Mining for 3D Retrieval}
%
% Single address.
% ---------------
\name{Hao WU, Ruochong LI, Hao Wang\sthanks{Corresponding author.}, Hui Xiong}
\address{The Hong Kong University of Science and Technology (Guangzhou), China}
\maketitle

\let\thefootnote\relax\footnotetext{This work was partially supported by Guangzhou-HKUST(GZ) Joint Funding Program (Grant No.2023A03J0008), Education Bureau of Guangzhou Municipality, Guangdong Science and Technology Department, and the Foshan HKUST Projects (FSUST21-FYTRI01A).}

\begin{abstract}
In this paper, we investigate an open research task of cross-modal retrieval between 3D shapes and textual descriptions. 
Previous approaches mainly rely on point cloud encoders for feature extraction, which may ignore key inherent features of 3D shapes, including depth, spatial hierarchy, geometric continuity, etc. 
To address this issue, we propose~\modelname, making the first attempt to exploit the cross-view correspondence and cross-modal mining to enhance the retrieval performance.
Notably, we augment the 3D features through a scene representation transformer, to generate cross-view correspondence features of 3D shapes, which enrich the inherent features and enhance their compatibility with text matching. 
Furthermore, we propose to optimize the cross-modal matching process based on the semi-hard negative example mining method, in an attempt to improve the learning efficiency. Extensive quantitative and qualitative experiments demonstrate the superiority of our proposed COM3D, achieving state-of-the-art results on the Text2Shape dataset. 
% Experiments
\end{abstract}
\vspace{-15pt}\section{Introduction}
\vspace{-10pt}Interactive 3D scenarios across entertainment, architecture, and automotive industries yield vast 3D, image, and text data. The integration of these cross-modal data representations is pivotal for enhanced comprehension and interaction with the three-dimensional aspects of the physical world. 
To facilitate a more intuitive interaction process, it is crucial to connect the realms of 3D data, image data, and linguistic data. Normally, 3D shapes and linguistic data can serve as complementary representations of object information. 
As the capabilities of Large Language Models (LLMs) advance rapidly, harnessing their robust capabilities to bridge the gap between textual descriptions and 3D features is increasingly critical. The task of aligning 3D shapes with corresponding textual information has emerged as a key challenge in the field.
% However, in numerous practical contexts, the 3D features characterized jointly by 3D shapes and linguistic data are neither complete nor unambiguous enough. This situation renders the amalgamation of 3D shapes, image data, and linguistic data into a cohesive framework an imperative endeavor.

Within this context, 3D retrieval emerges as one of the crucial tasks. Existing 3D retrieval methods~\cite{chang2015shapenet,tang2023parts2words,chen2019text2shape,han2019y2seq2seq,li2015joint,uy2021joint,yao2018mvsnet,sajjadi2022scene,fu2020hard,lin2021single} focus on learning a unified joint embedding space for linguistic data and other data types such as 3D shapes or images, employing metric or contrastive learning approaches for matching. However, in many practical scenarios, the 3D features jointly characterized by 3D shapes and linguistic data are often incomplete and ambiguous. This inadequacy can result in models suffering from poor local feature matching and slow computational speeds. It emphasizes the necessity to enhance the representation of 3D data in the joint embedding process. 

% \begin{figure}
% \centering   
%   \includegraphics[width=0.5\textwidth]{imgs/figure1.pdf}
% %   \vspace{-10pt}
% \caption{3D-to-Text retrieval results by our~\modelname. For each query sentence, we show the top-5 ranked shape.
%     }
%     \label{fig1:case}
%     \vspace{-10pt}
% \end{figure}

Another key aspect involves the matching of cross-modal embeddings using similarity metrics. While these techniques can be straightforward, the inherent gap between modalities like images or 3D data and text poses significant challenges. This disparity complicates the effective use of both global and local 3D semantic information, often resulting in poor matching performance and limited generalization. Meanwhile, the accuracy of conventional similarity measures in capturing cross-modal data nuances remains questionable. Consequently, devising an efficacious method to assess cross-modal similarity poses a significant challenge.

% Most existing methods focusing on two different types of data fail to fully utilize both global and local 3D semantic information, leading to subpar matching results and limited generalization. To address such issues, we propose a framework that fully utilizes information from different dimensions to represent the joint embedding space. For this space, we have designed appropriate similarity metrics. This approach is aimed at not only addressing the intricacies of multi-dimensional data interaction but also at establishing a robust framework for efficient and accurate 3D-text matching. Our model have achieved state-of-the-art (SOTA) results on the Past2word dataset.

%The challenge of 3D cross-modality retrieval lies in effectively embedding different data dimensions into a coherent joint representation space and designing appropriate similarity metrics within this space to achieve superior matching results and generalization performance.

% 少了scene representation transformer
In response to these improvement opportunities, we introduce a novel text-3D retrieval framework that leverages the multi-view correspondence features derived from point clouds. ~\modelname~ employs the Scene Representation transformer (SRT)~\cite{sajjadi2022scene} to extract features across different views using multi-view images, along with their associated camera poses and rays. The SRT facilitates the learning of joint 3D features from both point clouds and their multi-view images, thereby enriching the inherent 3D representation with aspects such as depth, spatial hierarchy, and geometric continuity. Additionally, we design a semi-hard negative mining method for our similarity metric, which combines the Earth Mover's Distance (EMD) with cosine similarity. 
% Overall, our approach culminates in an end-to-end model for 3D cross-modal retrieval. 
% joint text-image-3D matching method based on integrated embeddings to achieve the retrieval task of 3D retrieval, as depicted in Figure~\ref{fig0:model}. To optimize the matching metric in the unified representational space composed of these cross-modal data embeddings, we designed a loss function based on Earth Mover's Distance (EMD) and cosine similarity, thereby achieving the optimal matching between 3D and text data. Concurrently, we employed a semi-hard negative mining approach to refine the training process, enhancing the model's performance on confusable points and ambiguous edges. Overall, we propose an end-to-end 3D cross-modal retrieval model. 
\\
Our contributions can be summarized as:
\begin{itemize}
    \vspace{-7pt}\item We enhance our 3D shape representations by uncovering the semantic correspondence between multi-view point cloud images with the SRT transformer.
    \vspace{-7pt}\item We improve the text-3D retrieval learning efficacy and efficiency with the proposed semi-hard negative mining technique.
    \vspace{-7pt}\item We demonstrate our proposed \modelname~ framework outperforms several state-of-the-art methods for text-3D retrieval.
\end{itemize}

% \textbf{Our contributions.}   We propose a novel network framework for the joint embedding of textual, image, and 3D representations to facilitate their bidirectional matching. Utilizing [specific theory], we introduce a cosine similarity-based method for deriving rational Matching Scores among texts, images, and 3D shapes, which concurrently reduces computational load/increases computational speed. To the best of our knowledge, our model has achieved state-of-the-art (SOTA) results in the realms of joint 3D shape, image, and text understanding/retrieval.

\section{Related Work}
\vspace{-10pt}
\subsection{Joint Learning of Linguistic Data and 3D Shapes}
Recently, text-3D datasets such as ShapeNet~\cite{chang2015shapenet} and Parts2words~\cite{tang2023parts2words} enhance the joint learning of linguistic and 3D data, significantly advancing the geometric and semantic analysis in the 3D world. Text2shape~\cite{chen2019text2shape} introduces a framework based on joint embedding concepts to generate colored 3D shapes from linguistic data. Parts2words represents 3D shapes as point clouds and learns the joint embedding of point clouds and text while optimizing the feature space for enhanced matching, yet it overlooks the critical depth information and varying three-dimensional perspectives essential in text-3D tasks. Y$^2$seq2seq~\cite{han2019y2seq2seq} proposes a view-based model that bridges the semantic meanings embedded in both text and 3D data, focusing on global matching but neglecting the local features crucial to text-3D tasks. Tricolo~\cite{ruan2022tricolo} utilizes contrastive learning methods in joint representation learning for 3D and text, aiming to enhance the connection between these two modalities. 

PointCLIP and Clip2Point~\cite{Zhang_2022_CVPR,huang2023clip2point} transfer the ideas of CLIP~\cite{radford2021learning} and contrastive learning to 3D visual tasks, establishing joint embeddings of text, images, and point clouds based on rendered images. Uni3DL~\cite{li2023uni3dl} introduces a unified model for 3D and language understanding, which extends the range of 3D-supported tasks through direct manipulation of point clouds, facilitating a variety of 3D visual and language-related tasks.
%Each of these methods contributes uniquely to the evolving field of 3D and language representation learning, indicating a rich landscape of research and application possibilities.
\vspace{-10pt}
\subsection{Image-based 3D Shape Retrieval}
High-quality image-3D datasets like Pix3D~\cite{sun2018pix3d} and 3D-FUTURE~\cite{fu20213dfute} provide ample data support for the research of image-3D retrieval tasks. Typically, image-3D tasks aim to utilize a joint embedding space of 2D and 3D data and a similarity metric defined in this space to facilitate retrieval and other visual tasks~\cite{li2015joint,uy2021joint}. In this vein, multi-view Stereo (MVS) methods, such as MVSnet~\cite{yao2018mvsnet}, introduce an end-to-end deep learning model for depth inference to achieve image-3D matching, employing 3D convolutional networks. MVSTER~\cite{wang2022mvster} utilizes the epipolar Transformer to efficiently learn 2D semantics and 3D spatial associations, thereby reducing the network's running time. However, due to limitations in image clarity or tasks requiring high granularity, MVS methods normally performs below expectations.
%Following a similar conceptual framework, ET-MVSnet~\cite{Liu_2023_ICCV} enhances non-local feature matching based on epipolar geometry, while DMVSnet~\cite{Ye_2023_ICCV} considers depth geometric shapes in depth map prediction. However, due to limitations in image clarity or tasks requiring high granularity, MVS methods often underperform, prompting some works to focus on metric learning retrieval optimization~\cite{fu2020hard} and shape classification for high-granularity sketches~\cite{qi2021toward}. Yet, these methods still fall short in terms of generalization and speed.

Subsequently, SRT~\cite{sajjadi2022scene} emerges as a potential solution in these areas. The model is an end-to-end supervised system designed to minimize the reconstruction error of new 3D views, effectively integrating multi-view correspondence with 3D shapes to achieve exceptional speed and recognition accuracy. Building upon SRT, works like RUST~\cite{Sajjadi_2023_CVPR} have transcended the limitations of prior pose information and have expanded to handle complex real-world scenarios while maintaining quality performance.

%In the domain of 3D Retrieval, metric learning-based methods~\cite{fu2020hard} often map 2D images and 3D shapes into an embedding space and learn a similarity metric for matching. Patch2cad~\cite{kuo2021patch2cad} employs object view segmentation for similarity matching between 2D images and 3D CAD models. Additionally, contrastive learning methods~\cite{lin2021single} have found applications in image-based 3D shape retrieval, further pushing the boundaries of precision and effectiveness in this evolving field. Each model and method, with its unique approach and application, continues to contribute to the refinement and advancement of image-3D retrieval tasks.
\label{sec:relat}

\label{sec:intro}

\begin{figure*}
\centering   
  \includegraphics[width=1\textwidth]{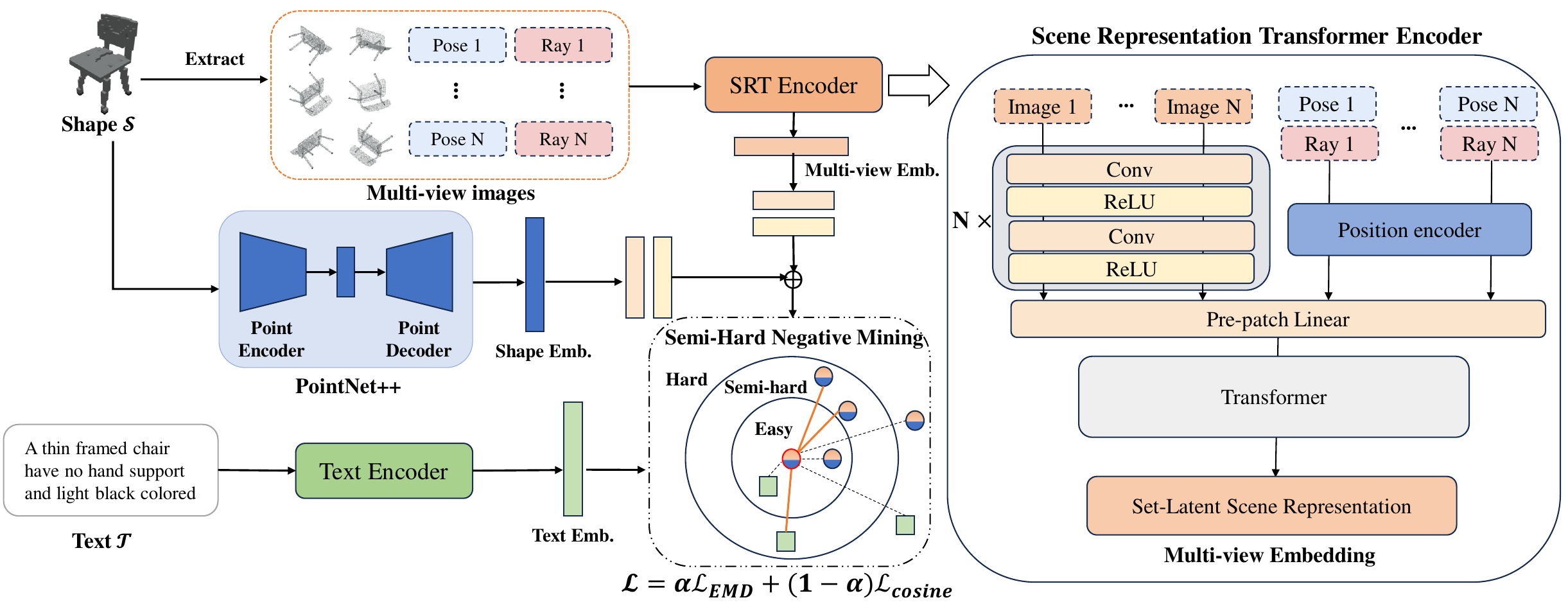}
  \vspace{-10pt}
\caption{An overview of our proposed~\modelname. It is composed of three main embeddings: \textbf{multi-view embedding}, \textbf{shape embedding}, and \textbf{text embedding}. The multi-view embedding is derived from multi-view images using a \textbf{Scene Representation Transformer (SRT) Encoder}, considering the corresponding camera poses and rays; Shape embedding is extracted by PointNet++. \textbf{Semi-Hard Negative Mining} enhances the matching of text and 3D by focusing on moderately challenging samples within the semi-hard range, marked by \textcolor{orange}{orange lines}, with anchor depicted by \textcolor{red}{red borders}.
    }
    \label{fig0:model}
    \vspace{-10pt}
\end{figure*}

\section{Method}
\textbf{Overview} As shown in Fig.~\ref{fig0:model}, the~\modelname~ represents an innovative framework designed for the 3D retrieval task, comprising two essential modules: a joint encoder module and a matching module. To construct a unified joint embedding representation space for matching features of 3D shapes, multi-view images, and text, we initially apply an encoder
%based on Scene Representation Transformer (SRT)
to embed the multi-view images from 3D shapes. Concurrently, a joint embedding is accomplished by combining a 3D shape encoder
%based on Point-Net++ 
and a text encoder,
%based on Bi-directional Gate Recurrent Unit (GRU)
thereby achieving joint embedding across modalities.

To optimize the matching between the fusion embedding$~\mathcal{F}_{\text{fused}}$ which comprises 3D shapes, multi-view images, and text embeddings$~\mathcal{T}$, we introduce a matching module to score the 3D shapes and linguistic data. After that, we utilize a matching loss composed of Earth Mover's Distance (EMD) and Cosine similarity, training our model based on the semi-hard negative mining method.

%This architecture encompasses three modules: two 3D encoders for 3D shape and multi-view feature extraction, a text encoder for textual feature extraction, and a retrieval module for matching the text and the 3D shape features.

%We first discuss the 3D retrieval task by defining its input. Our \modelname~ takes as input $\mathcal{P}$ from the point cloud dataset.
%
%\begin{itemize}
%\item Point cloud data $\mathcal{P}$
%\item Text $\mathcal{S}$
%\item multi-view embedding $\mathcal{M}$
%\end{itemize}

\subsection{Preliminary}

\subsubsection{The Point-cloud Encoder}
Point-cloud encoder such as PointNet++~\cite{qi2017pointnet++} aims to extract features from point cloud data \( \mathcal{P} \). The feature propagation in PointNet++ is achieved through a hierarchical process, interpolating features across varying resolutions of the point set. For a given set of points \( \mathcal{P} \), with each point \( x_i \) having its associated feature vector \( f_i \), the interpolated feature \( \hat{f_i} \) at each point is computed by a distance-based weighted average of the features from its \( k \) nearest neighbors.

The feature propagation is formalized as follows:
\begin{equation}
\hspace{-1em}\hat{f_i}(x) = \frac{\sum_{j=1}^{k} w_{ij}(x) f_{j}(x)}{\sum_{j=1}^{k} w_{ij}(x)}, \quad \hspace{-0.7em} \text{where\hspace{-0.7em}} \quad w_{ij}(x) = \frac{1}{d(x, x_j)^p}
\end{equation}
and \( d(x, x_j) \) denotes the Euclidean distance between the points \( x \) and \( x_j \), with \( p \) controlling the influence of the distance on the weighting function. This process iteratively updates the feature vectors for each point in \( \mathcal{P} \), yet yielding a set of enriched feature vectors \( \mathcal{F} = \{ f_i \} \) that capture the complex structure of the point cloud.

\vspace{-10pt}
\subsubsection{The Text Encoder}
\vspace{-5pt}
Given a text sequence or caption \( \mathcal{S} \), consisting of \( m \) words, we define the embedding of the \( t \)-th word as \( \mathbf{e}_t \). The text encoder computes the hidden states for each word using a bidirectional GRU, resulting in a sequence of word embeddings:
\vspace{-5pt}
\begin{equation}
\begin{aligned}
h_t^f &= \text{GRU}^f(e_t, h_{t-1}^f), & t &\in [1, m] \\
h_t^r &= \text{GRU}^r(e_t, h_{t+1}^r), & t &\in [m, 1] \\
w_t &= \frac{h_t^f + h_t^r}{2}, & t &\in [1, m]
\end{aligned}
\end{equation}

Where$~\mathbf{h}_t^f$ is the forward hidden state of the GRU for the \( t \)-th word.$~\mathbf{h}_t^r $ is the reverse hidden state of the GRU for the \( t \)-th word.$~\mathbf{w}_t$ is the final embedding for the \( t \)-th word, obtained by averaging its corresponding forward and reverse hidden states.
These embeddings $~\{\mathbf{w}_t\}_{t=1}^m$ are then used to represent the text \( \mathcal{S} \) within our model.

\vspace{-10pt}
\subsection{The Cross-modal Joint Encoder}
\vspace{-5pt}
\subsubsection{Scene Representation Transformer Encoder}
\vspace{-5pt}
The Scene Representation Transformer (SRT)~\cite{sajjadi2022scene} was developed to quickly infer 3D scene representations from posed or unposed images for rendering novel views.
Given the point cloud data \( \mathcal{P} \), our model first extracts a set of multi-view images \( \mathcal{I} = \{I_i\}_{i=1}^{V} \), where \( V \) is the number of views, and each image \( I_i \) is associated with a specific camera pose \( c_i \) and ray \( r_i \) from the set \( \mathcal{C} \) and \( \mathcal{R} \), respectively.

In the SRT, image data and associated ray information are initially transformed into feature vectors through a RayEncoder that employs positional encoding. This encoding process incorporates multiple octaves, beginning at a designated starting octave, which is then utilized to generate an encoded feature vector $x$ for each image. Image features are processed through convolutional layers, doubling dimensions until a set size, and transformed into patch embeddings. These are enhanced with positional data and camera-specific characteristics, then unified by a transformer architecture into a comprehensive 3D scene representation. The mathematical representation of the SRT encoder can be formalized as:
\vspace{-5pt}
\begin{equation}
    \mathcal{M} = Trans\left(\text{Concat}\left(\mathcal{I}_{\text{conv}} + \mathcal{E}_{\text{pos}}, \mathcal{C}_{\text{emb}}\right)\right)
\end{equation}
where \( \mathcal{I}_{\text{conv}} \) denotes the image features after convolutional processing, \( \mathcal{E}_{\text{pos}} \) is the positional embedding for pixels, \( \mathcal{C}_{\text{emb}} \) is the camera position embedding, and $Trans$ represents the transformer module.
%\vspace{-20pt}
\subsubsection{Cross-modal Joint Encoder}
\vspace{-5pt}
Leveraging the information from multiple distinct modalities can fully exploit the characteristics of three dimensions, optimizing subsequent 3D-Text matching. This necessitates the construction of an appropriate unified joint embedding space to thoroughly fuse the features of 3D shapes, multi-view images, and text. Consequently, we propose a Cross-modal Joint Encoder, composed of encoders corresponding to three different types of information: a 3D shape encoder, a multi-view encoder, and a text encoder. These encoders embed the data into the same representational space, setting the stage for 3D Retrieval.

\subsection{The Retrieval Module}
\vspace{-5pt}
\subsubsection{Optimal Matching Loss Function}
In our retrieval module, we adopt the EMD to assess the alignment between 3D shape embeddings \( \mathcal{F} \) and textual embeddings \( \mathcal{T} \). Additionally, we integrate a cosine similarity loss to enhance the retrieval accuracy. The 3D shape embeddings are first merged with the multi-view embeddings \( \mathcal{M} \) through a transformation, which consists of a fully connected layer and a subsequent ReLU activation function. This results in a fused embedding \( \mathcal{F}_{\text{fused}} \).
%which is mathematically expressed as:
%\begin{equation}
%\mathcal{F}_{\text{fused}} = \text{$ReLU$}(\text{FC}(\mathcal{F}, \mathcal{M}))
%\end{equation}

After obtaining \( \mathcal{F}_{\text{fused}} \), the cosine similarity $sim$ with the text embeddings \( \mathcal{T} \) is calculated, followed by the application of the EMD framework to optimize the matching flow between the shape and text representations. The cosine similarity is explicitly incorporated into the EMD optimization as an additional term in the loss function, guiding the model towards embeddings that are not only transport-efficient but also directionally aligned in the embedding space. The combined loss is represented as:
\vspace{-5pt}
\begin{equation}
\hspace{-1em}\mathcal{L}_{\text{r}} = \text{$EMD$}(\mathcal{F}_{\text{fused}}, \mathcal{T}) - \lambda \sum_{i=1}^{n}\sum_{j=1}^{m} \text{$sim$}(\mathcal{F}_{\text{fused}_i}, \mathcal{T}_j) \cdot \hat{x}_{ij}
\end{equation}

 where \( \lambda \) is a balancing coefficient, \( \hat{x}_{ij} \) denotes the optimal transport plan computed via the Sinkhorn algorithm, and \( \text{$sim$}(\mathcal{F}_{\text{fused}_i}, \mathcal{T}_j) \) is the cosine similarity between the fused shape embedding \( \mathcal{F}_{\text{fused}_i} \) and the text embedding \( \mathcal{T}_j \).
 \vspace{-5pt}
\subsubsection{Semi-hard Negative Mining in Retrieval}
\vspace{-5pt}
Semi-hard negative mining is a strategy used in machine learning, especially in metric learning and deep learning tasks, aiming to improve the discriminative power of models. It involves selecting negative samples that are neither too easy nor too hard for the model to classify — these are referred as semi-hard negatives. This technique is crucial in the context of triplet loss, where the goal is to ensure that an anchor is closer to positive samples than to negative samples in the feature space. 

Utilizing semi-hard negative mining within the representation space based on 3D shapes, multi-view images, and text joint embedding not only enhances the matching effectiveness of text but also facilitates the model's ability to learn more robust and generalizable features. By focusing on semi-hard negatives—those that are neither the easiest nor the most challenging to distinguish—the method demonstrates exceptional performance in learning difficult-to-classify target points and ambiguous local boundary features. This focus leads to improved performance in tasks that require fine-grained discrimination between similar classes. In our work, semi-hard negative mining is applied during the training stage, where it directly influences the loss functions of EMD and Cosine similarity, integral to the model's learning process. Taking EMD based mining as an example, the steps typically involved in applying semi-hard negative mining are as follows:

Select Semi-hard negatives:
Select negative samples that are far enough from the positive sample but not the furthest. This is done by comparing each negative sample's EMD score to the positive sample's EMD score, selecting those with higher but not the highest scores as semi-hard negatives.

Apply Margin:
With semi-hard negatives selected, a predefined margin is used for pattern calculation adjustment. Calculation formula:
\begin{equation}
\mathcal{L}_{\text{semi-hard}} = \max(0, \text{$margin$} + \text{$score$}_{\text{$neg$}} - \text{$score$}_{\text{$pos$}})
\end{equation}
where \( \text{$score$}_{\text{$pos$}} \) is the positive sample's EMD score and \(\text{$score$}_{\text{$neg$}} \) is the semi-hard negative's EMD score.
By minimizing this loss, the model learns to distinguish semi-hard negative samples from positive samples, thus improving its ability to differentiate between different types of sample categories.

\begin{table*}
\centering
\begin{tabular}{l|ccc|ccc}
\toprule
\textbf{Method} & \multicolumn{3}{c}{\textbf{S2T}} & \multicolumn{3}{c}{\textbf{T2S}} \\
 & \textbf{RR@1} & \textbf{RR@5} & \textbf{NDCG@5} & \textbf{RR@1} & \textbf{RR@5} & \textbf{NDCG@5} \\ \midrule
Text2Shape~\cite{chen2019text2shape}  & 0.83 & 3.37 & 0.73 & 0.40 & 2.37 & 1.35 \\
Y\textsuperscript{2}Seq2Seq~\cite{han2019y2seq2seq} & 6.77 & 19.30 & 5.30 & 2.93 & 9.23 & 6.05 \\
TriCoLo~\cite{ruan2022tricolo} & 16.33 & 45.52 & 12.73 & 10.25 & 29.07 & 19.85 \\
Parts2Words~\cite{tang2023parts2words} & 19.38 & 47.17 & 15.30 & 12.72 & 32.98 & 23.13 \\\midrule
~\modelname & \textbf{20.03} & \textbf{48.32} & \textbf{15.62} &\textbf{13.12} &\textbf{33.48} & \textbf{23.89}\\
\bottomrule

\end{tabular}
\caption{Comparison of different methods on S2T and T2S tasks. The results uses \textbf{additional semantic part segmentation} data in Parts2Words~\cite{tang2023parts2words}.}
\vspace{-15pt}
\label{tab:result_1}
\end{table*}
\vspace{-15pt}
\subsection{Overall Objective}
\vspace{-5pt}
The overarching goal of \modelname\ is to effectively retrieve 3D shapes by matching them with corresponding textual descriptions. To achieve this, we employ a comprehensive approach that involves feature extraction from both the geometric and visual domains, as well as from the textual domain. Our method synthesizes these multimodal inputs into a cohesive representation that is conducive to accurate retrieval.

The joint embeddings from the 3D shapes and multi-view images are processed through a sequence of transformational layers to produce fused embeddings \( \mathcal{F}_{\text{fused}} \). Concurrently, text embedding \( \mathcal{T} \) are extracted and prepared for comparison. The retrieval process is framed as an optimization problem that leverages the Earth Mover's Distance (EMD) while incorporating a cosine similarity metric to refine the matching between 3D shapes and texts.

The final loss function encapsulates both the transport efficiency and the alignment in the feature space, formulated as a weighted sum of the EMD and the cosine similarity losses. It is given by:
\vspace{-5pt}
\begin{equation}
\mathcal{L}_{\text{r}} = \alpha \mathcal{L}_{\text{EMD}} + (1 - \alpha) \mathcal{L}_{\text{cos}}
\end{equation}

where \( \alpha \) is a hyperparameter that trades off the contribution of the retrieval loss \( \mathcal{L}_{\text{EMD}} \) and the cosine similarity loss \( \mathcal{L}_{\text{cos}} \). This combined loss function is minimized during training to ensure that the resulting embeddings for the 3D shapes and the textual descriptions are closely aligned, facilitating accurate retrieval in response to textual queries.

\begin{figure}
\centering   
  \includegraphics[width=0.5\textwidth]{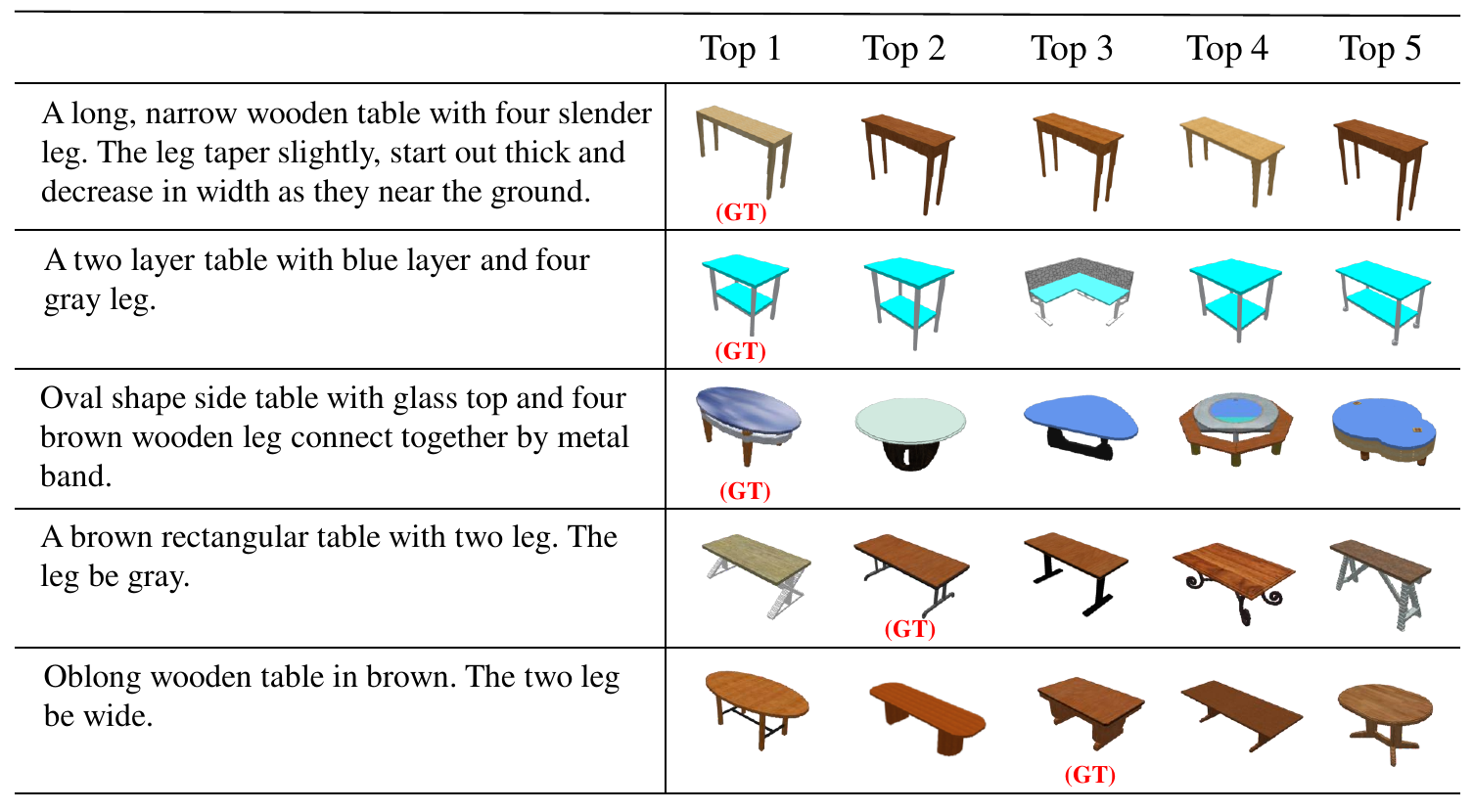}
  \vspace{-10pt}
\caption{text-to-3D retrieval results by our~\modelname. For each query sentence, we show the top-5 ranked shape.
    }
    \label{fig1:case}
    \vspace{-10pt}
\end{figure}
\vspace{-10pt}
\section{Experiments}
\subsection{Experimental settings}
\vspace{-5pt}
\noindent \textbf{Data Preparation}
In the 3D retrieval task, we employ a dataset~\cite{tang2023parts2words} that provides both 3D representations and corresponding textual descriptions to enable a comprehensive evaluation of our framework. We utilize the well-constructed 3D-Text cross-modal dataset, which is specifically designed to bridge the modalities of 3D shapes and natural language descriptions.

This dataset is split into a training set consisting of 11,498 3D shapes, and a test set with 1,434 shapes. Each shape is accompanied by an average of five textual descriptions, enabling the model to learn and predict from multiple descriptive angles.
It stands out due to its utilization of detailed 3D models from ShapeNet~\cite{chang2015shapenet} and PartNet~\cite{mo2019partnet}. ShapeNet provides the fundamental 3D shapes, while PartNet offers a more intricate layer of fine-grained, instance-level segmentation annotations. Such a combination allows for a rich representation of objects that our model can leverage.

\noindent \textbf{Evaluation metrics}
For the evaluation of our 3D retrieval model, we adopt the standard metrics of Recall Rate at $k$ (RR@$k$), specifically for $k = 1$ and $k = 5$, and the Normalized Discounted Cumulative Gain (NDCG). 
RR@$k$ quantifies the likelihood of the correct match appearing within the top $k$ retrieved results. NDCG assesses the ranking quality of the retrieval process, taking into account the position of relevant items within the retrieval list. 

\noindent \textbf{Parameter Setting}
we configure the shape encoder to process point clouds fixed at 2500 points each, segmented according to a granular scheme of 17 classes. We employed PointNet++~\cite{qi2017pointnet++} architectures within our shape encoder module to evaluate their performance in feature extraction. The aggregation of features was facilitated through average pooling in our group pooling module, deliberately ignoring segments constituting less than 1\% of the point cloud to prioritize salient features.
For the 3D retrieval task, embeddings were standardized to a 1024-dimensional space. Textual encoding was performed using a single-layer Bi-directional GRU, deliberately trained from scratch, owing to negligible vocabulary benefits observed from pre-training techniques. The optimization of our model employed a semi-hard negative mining strategy within its triplet ranking loss, with an established margin of 0.2.
An Adam optimizer guided the learning process, initiated with a learning rate of 0.001.

% The training regimen was bifurcated into an initial phase focusing solely on semantic segmentation for 50 epochs, followed by a multitask learning phase over 20 epochs, balancing the losses with a weight factor of 40.
\vspace{-10pt}
\subsection{Comparison results}
We evaluate our framework in the dataset from part2words~\cite{tang2023parts2words}, which includes additional semantic part segmentation data for the 3D point cloud. Accordingly, our comparison results include scenarios both with and without part segmentation. We compare with many previous text-3D retrievals methods such as Text2Shape~\cite{chen2019text2shape}, Y2seq2seq~\cite{han2019y2seq2seq}, and TriCoLo~\cite{ruan2022tricolo}. our framework showed consistent superiority and achieved state-of-the-art (SOTA) performance in various settings.

Part2words showcases specific optimizations for part segmentation features. In contrast, our framework enhances 3D semantic features through the inherent characteristics of multi-view images in point clouds facilitated by the Scene Representation Transformer (SRT). This strategy led to a marked improvement in performance, enabling our method to surpass part2words and achieve state-of-the-art (SOTA) results, as detailed in Tab.~\ref{tab:result_1}. 

Our method outperforms previous approaches in experiments without part segmentation, as shown in Tab.~\ref{tab:result_2}. This outcome underscores the remarkable generalization capability of our model, evidencing the efficacy of the SRT optimization across various settings. Our model consistently delivers SOTA performance in different configurations, showcasing its robustness and wide applicability. Besides, the visualization results are shown in Fig.~\ref{fig1:case}.

\vspace{-5pt}
\subsection{Ablation study}
\noindent \textbf{EMD and cosine loss}
We evaluate the impact of the Earth Mover's Distance (EMD) and cosine loss on our retrieval module, as detailed in Tab.~\ref{tab:result_2}. In our ablation analysis with \modelname, we observe a noticeable performance decline when either of the losses is removed. Notably, the absence of EMD results in a more pronounced drop in performance. This can be attributed to EMD's focus on optimizing local details, while cosine similarity concentrates on the overall alignment in the vector space. These findings underscore the synergistic importance of both losses in enhancing the retrieval process.

\noindent \textbf{SRT}
The SRT is a cornerstone of our approach, extracting multi-view features from point clouds. Compared with our baseline, the addition of SRT significantly boosts performance. This improvement is attributed to the ability of SRT to utilize information overlooked by PointNet++ and to enhance the matching process with text from the multi-image perspective.

\noindent \textbf{Semi-hard negative mining}
Semi-hard negative mining is an optimization technique that we analyze in contrast to the commonly used hardest negative mining. Replacing semi-hard with the hardest negative mining leads to a sharp decline in performance. This may be due to the unsuitability of the hardest negative mining for retrieval tasks and the specific variable characteristics of our model. The effectiveness of semi-hard negative mining is thus validated by these comparative results.
% 1. 主要实验提升，
%表一是我们在相关方法上取得的量化结果的展示。在Parts2words数据集上，我们的模型比别的方法取得了更好的效果。在xx指标上取得了【】的提升,展现出了SOTA的结果。

%主要是对比part2words，分两部分，1是带part segment的，2是不带part segment（参考一下part2word 这个到底叫啥）的，不带part segment的效果更明显，分析一下原因是我们的模型更通用，没有针对segment做优化，但是在不同setting都有提升。可以细讲一下这2个不同任务和6个指标的意义
% The main result 主实验提升

% ablation study 1. 验证SRT 两个loss的有效性 EMD和cosin loss对我们的模型都有提升，相似度，编一些理由 

%2. SRT的有效性，删去SRT后性能下降明显，说明我们的SRT对3D feature的提取

%3. semi-hard mining的效果也比较明显。 我们对比了 hardest negative mining and semi-hard negative mining，发现hardest negative mining效果很差，不适合这个任务，。。。

% visualization 可视化对比，

\begin{table}
\centering
\renewcommand{\tabcolsep}{3mm}
  \resizebox{\linewidth}{!}{
\begin{tabular}{l|ccc}
\toprule
\textbf{Method} &  \multicolumn{3}{c}{\textbf{T2S}} \\
&\textbf{RR@1} & \textbf{RR@5} & \textbf{NDCG@5} \\ \midrule
Text2Shape & 0.40 & 2.37 & 1.35 \\
Y\textsuperscript{2}Seq2Seq & 2.93 & 9.23 & 6.05\\
Part2Words  & 5.06 & 17.21 & 11.25 \\ \hline
\modelname & \textbf{5.64} & \textbf{18.50} & \textbf{12.09}\\
~~- cosine loss & 5.12 & 17.20 & 11.54 \\
~~- EMD loss & 4.83 & 16.32 & 10.42\\
~~- SRT & 5.07 & 17.92 & 11.52 \\
~~- semi-hard & 0.93 & 1.92 & 1.42\\

\bottomrule

\end{tabular}}
\caption{Comparison of different methods on T2S tasks and the \textbf{ablation study} on our \modelname. The results use the Text2Shape\cite{chen2019text2shape} ShapeNet subset \textbf{without semantic part segmentation}. }
\vspace{-10pt}
\label{tab:result_2}
\end{table}

\vspace{-10pt}
\section{Conclusion}
\vspace{-10pt}
In this paper, we introduce an efficient end-to-end cross-modal 3D retrieval model. This model utilizes cross-modal data to achieve bidirectional matching between different types, implementing text-3D retrieval. We initially propose a cross-modal joint embedding model to fully integrate the features of text, multi-view images, and 3D shapes. Moreover, we measure the features across different dimensions using cosine similarity in the joint embedding space and employ a hard sample mining approach to achieve improved matching results. Our findings have achieved state-of-the-art (SOTA) results on the past2word dataset. One limitation of our model is that it only considers point cloud features due to the lack of sufficient datasets. However, the concept of joint learning and matching based on different types of data is highly efficient, endowing our model with the potential to extend to various 3D representations such as mesh.
\vspace{-5pt}

\bibliographystyle{IEEEbib}
% \footnotesize
\bibliography{icme2023template}

\input{supp}
\end{document}

%% file: supp.tex
\appendix
\renewcommand{\thesection}{\Alph{section}}

\section{More experiment result}
\begin{figure*}
\centering   
  \includegraphics[width=1\textwidth]{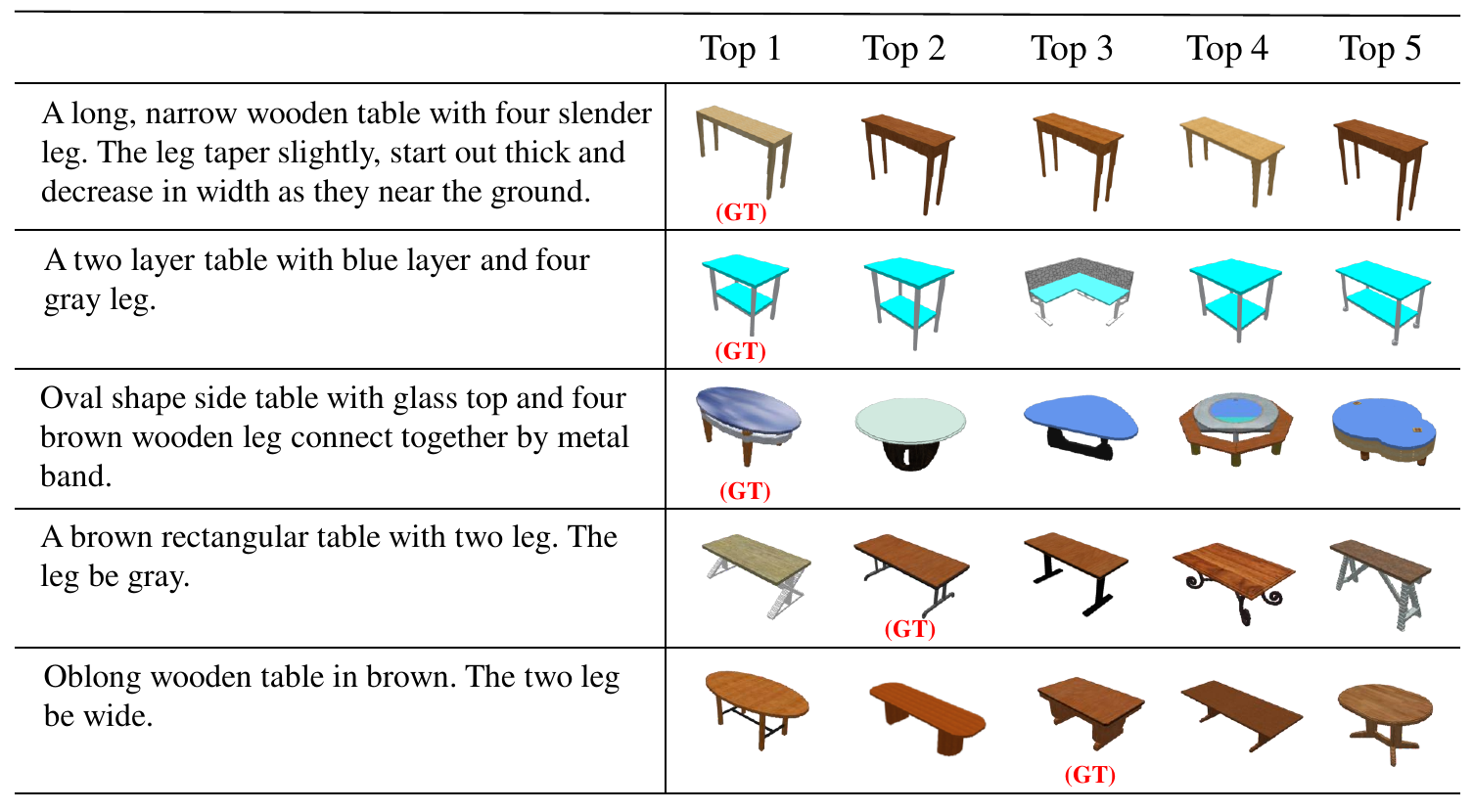}
\caption{text-to-3D retrieval results by our~\modelname. For each query sentence, we show the top-5 ranked shape.
    }
    \label{fig3:case}
\end{figure*}

\begin{figure*}
\centering   
  \includegraphics[width=0.7\textwidth]{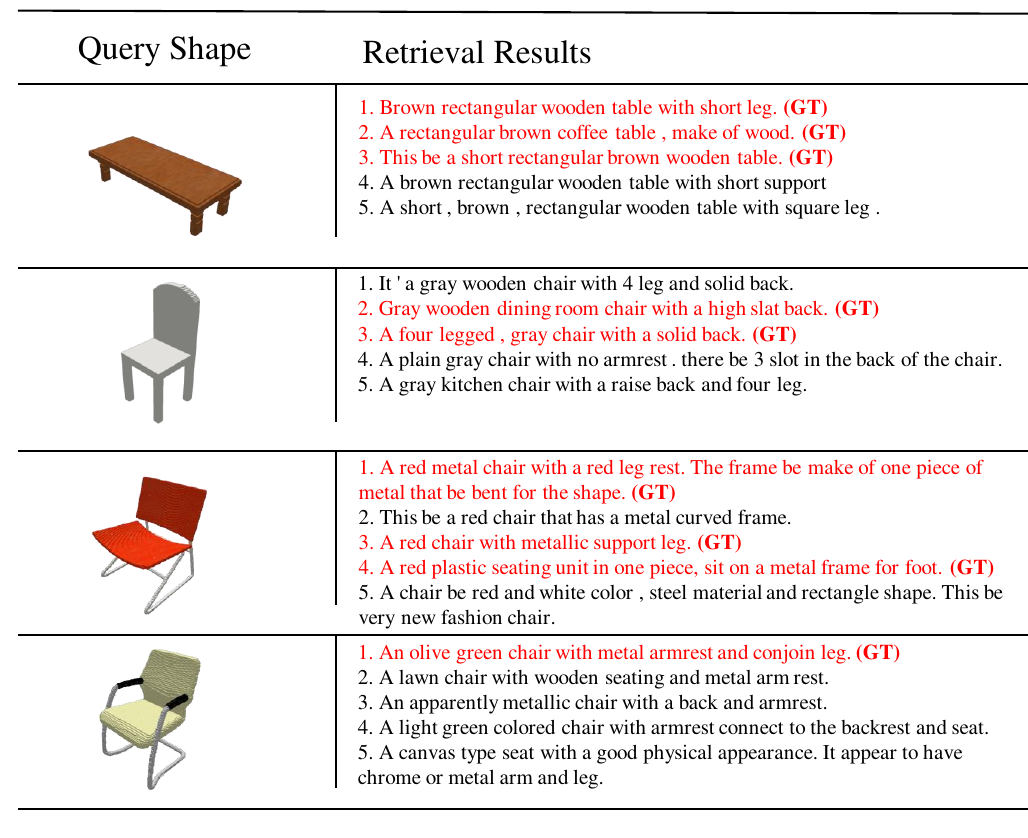}
\caption{3D-to-text retrieval results by our~\modelname. For each query sentence, we show the top-5 ranked shape.
    }
    \label{fig2:case}
\end{figure*}

More visualization results of our ~\modelname~ are shown in Fig.~\ref{fig3:case} and Fig.~\ref{fig2:case}.

\section{More details of the proposed method}
\subsection{The Cross-modal Joint Encoder}
\subsubsection{Scene Representation Transformer Based Encoder}
The Scene Representation Transformer (SRT)~\cite{sajjadi2022scene} has been innovatively adapted to process 3D scenes from various perspectives, enabling novel view rendering. In our framework, this transformer plays a pivotal role in encoding multi-view images derived from 3D point clouds. For a set of point cloud data \( \mathcal{P} \), we extract a collection of multi-view images \( \mathcal{I} = \{I_i\}_{i=1}^{V} \), where \( V \) represents the total number of views. Each image \( I_i \) is linked with a distinct camera pose \( c_i \) and ray \( r_i \).

The SRT encoder operates in the following stages:
\begin{enumerate}
    \item \textit{Ray Encoding}: Each image and its associated ray information are encoded using a RayEncoder, applying positional encoding to create a feature vector \( x \) for each image.
    \item \textit{Convolutional Processing}: The encoded image features undergo a series of convolutional transformations, with the hidden dimension successively doubling until reaching the target size.
    \item \textit{Patch Embedding}: Post-convolutional features are reshaped into patch embeddings and combined with positional embeddings and camera-specific embeddings.
    \item \textit{Transformation}: A transformer architecture integrates all patch embeddings, resulting in a comprehensive representation of the 3D scene.
\end{enumerate}

Mathematically, the SRT encoder is expressed as:
\begin{equation}
\begin{split}
    \mathcal{M} = \text{Transformer}\Big(&\text{Concat}\big(\text{Conv}(\mathcal{I}) \\
    &+ \text{PosEmbed}(\mathcal{I}), \text{CamEmbed}(\mathcal{C})\big)\Big)
\end{split}
\end{equation}

where \( \text{Conv}(\mathcal{I}) \) denotes convolutional features of the images, \( \text{PosEmbed}(\mathcal{I}) \) represents positional embeddings, and \( \text{CamEmbed}(\mathcal{C}) \) corresponds to the camera position embeddings. The transformer module is denoted as \( \text{Transformer} \).

\subsubsection{Cross-modal Joint Encoder}
To effectively harness the characteristics of 3D data, we propose a Cross-modal Joint Encoder, integrating encoders for 3D shape, multi-view images, and text into a unified embedding space. Our approach consists of:
\begin{enumerate}
    \item \textit{3D Shape Encoder}: Utilizing PointNet++~\cite{qi2017pointnet++} for extracting features from point clouds.
    \item \textit{Multi-view Encoder}: Based on the SRT, it synthesizes diverse views of 3D shapes into a single scene representation.
    \item \textit{Text Encoder}: Employs a Bi-directional GRU for bidirectional text feature extraction.
\end{enumerate}
These encoders collectively map their respective inputs into a harmonized feature space, facilitating efficient 3D-text retrieval.

\subsection{The Retrieval Module}
\subsubsection{Optimal Matching Loss Function}
Our retrieval module combines EMD and cosine similarity to optimize the matching process.